\documentclass{article}

\usepackage{arxiv}

\usepackage[utf8]{inputenc} 
\usepackage[T1]{fontenc}    
\usepackage{hyperref}       
\usepackage{url}            
\usepackage{booktabs}       
\usepackage{amsfonts}       
\usepackage{nicefrac}       
\usepackage{microtype}      
\usepackage{lipsum}		
\usepackage{graphicx}
\usepackage{natbib}
\usepackage{doi}

\usepackage{url,hyperref,lineno,microtype,subcaption}
\usepackage[onehalfspacing]{setspace}

\usepackage{cite}
\usepackage{amsmath,amssymb,amsfonts}
\usepackage{algorithmic}
\usepackage{graphicx}
\usepackage{textcomp}
\usepackage{subcaption}
\usepackage{cases}
\usepackage{hyperref}

\usepackage{amsthm}
\usepackage{cite}
\theoremstyle{remark}
\newtheorem{remark}{Remark}

\usepackage{diagbox}
\usepackage[font=footnotesize]{caption}
\newcommand{\norm}[1]{\left\lVert#1\right\rVert}

\title{Mediating between Contact Feasibility and Robustness of Trajectory Optimization through Chance Complementarity Constraints}

\author{ \href{https://orcid.org/0000-0002-9908-8641}{\includegraphics[scale=0.06]{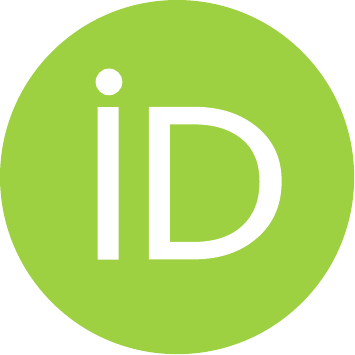}\hspace{1mm}Luke Drnach}\\
	Department of Electrical and Computer Engineering\\
	Georgia Institute of Technology\\
	Atlanta, GA 30318 \\
	\texttt{luke.drnach@gatech.edu} \\
	\And
	\href{https://orcid.org/0000-0002-7442-5920}{\includegraphics[scale=0.06]{orcid.pdf}\hspace{1mm}John Z. Zhang} \\
	George W. Woodruff School of Mechanical Engineering\\
    Georgia Institute of Technology\\
	Atlanta, GA 30318 \\
	\texttt{zzhang741@gatech.edu} \\
	\And
	\href{https://orcid.org/0000-0000-0000-0000}{\includegraphics[scale=0.06]{orcid.pdf}\hspace{1mm}Ye Zhao} \\
	George W. Woodruff School of Mechanical Engineering\\
    Georgia Institute of Technology\\
	Atlanta, GA 30318 \\
	\texttt{ye.zhao@me.gatech.edu} \\

}

\renewcommand{\shorttitle}{Mediating Contact Feasibility and Robustness}

\hypersetup{
pdftitle={Mediating between Contact Feasibility and Robustness of Trajectory Optimization through Chance Complementarity Constraints},
pdfsubject={cs.RO},
pdfauthor={Luke Drnach, John Z.~Zhang, Ye Zhao},
pdfkeywords={chance constraints, complementarity constraints, planning with contact, robust motion planning, trajectory optimization},
}

\begin{document}
\maketitle

\begin{abstract}
As robots move from the laboratory into the real world, motion planning will need to account for model uncertainty and risk. For robot motions involving intermittent contact, planning for uncertainty in contact is especially important, as failure to successfully make and maintain contact can be catastrophic. Here, we model uncertainty in terrain geometry and friction characteristics, and combine a risk-sensitive objective with chance constraints to provide a trade-off between robustness to uncertainty and constraint satisfaction with an arbitrarily high feasibility guarantee. We evaluate our approach in two simple examples: a push-block system for benchmarking and a single-legged hopper. We demonstrate that chance constraints alone produce trajectories similar to those produced using strict complementarity constraints; however, when equipped with a robust objective, we show the chance constraints can mediate a trade-off between robustness to uncertainty and strict constraint satisfaction. Thus, our study may represent an important step towards reasoning about contact uncertainty in motion planning.
\end{abstract}


\section{Introduction}

\label{sec:introduction}
As robots move into the real world, accounting for model uncertainty and risk in motion planning will become increasingly important. While model-based planning and control has demonstrated success in designing and executing dynamic motion plans for robots in a variety of tasks in the laboratory \citep{dai2014whole, Mordatch12, winkler_gait_2018,patel_contact-implicit_2019}, real world environments are difficult or intractable to precisely model, and as such the resulting motion plans could be prone to failure due to modeling errors. Planning for uncertainty and risk is especially important when the task involves intermittent contact, as incorrectly modeling friction can cause robots to drop and break objects or slip and fall, and incorrectly modeling contact geometry can cause mobile robots to trip and fall or collide with obstacles. While decent controller design can mitigate the effects of small modeling errors and disturbances \citep{toussaint_dual_2014, gazar_stochastic_2020}, incorporating uncertainty and risk into planning can help improve performance by generating reference trajectories that have a high success rate for execution.

Trajectory optimization is powerful for planning continuous dynamic motions that obey constraints such as actuation limits, obstacle avoidance, and contact dynamics \citep{posa2014direct, dai_optimizing_2012, dai2014whole,carius_trajectory_2018, gazar_stochastic_2020, Kuindersma16, mordatch2015ensemble, yeganegi_robust_2019}. While the optimal strategies produced by trajectory optimization typically lie on the boundary of the feasible region, recent works have begun to incorporate risk and uncertainty to improve the robustness of the planned motion. Uncertainty about the state or dynamics can be accounted for by an expected exponential transformation of the cost, resulting in risk-sensitive trajectory optimization \citep{ponton_role_2018, farshidian_risk_2015}. Alternatively, uncertainty about the constraints has been approached by defining failure probabilities and optimizing for motion plans that do not exceed some user-defined total failure probability \citep{hackett_risk-constrained_2020, shirai_risk-aware_2020}. Planning under contact uncertainty, however, has only recently begun to be investigated. One recent work developed a risk-sensitive cost term to plan for uncertainty in the contact model for systems with intermittent contact \citep{drnach_robust_2021}. However, while the robust cost formulation for uncertainty in contact produced robust trajectories, it also produced infeasible motion plans at high uncertainty, including setting friction forces to zero during sliding and allowing for positive contact reactions at nonzero contact distance. 

In this work, we explicitly investigate uncertainty resulting from the terrain contact parameters and develop a method for trading off between motion feasibility and robustness. In contrast to the previous work \citep{drnach_robust_2021}, which controlled robustness only by varying the uncertainty, we aim to achieve a tradeoff at fixed uncertainty by introducing tunable risk parameters.  Specifically, we: 
\begin{itemize}
    \item Design chance constraints for contact with uncertainty in contact distance and friction coefficient.
    \item Provide a risk-bounded interpretation to the relaxed chance complementarity constraints.
    \item Demonstrate that chance constraints, combined with a contact-sensitive objective, can control the trade-off between robustness to contact uncertainty and contact constraint satisfaction at fixed values of uncertainty. 
\end{itemize}

\section{Related Work}
\subsection{Contact-Robust Trajectory Optimization}
Planning motions for robots with intermittent contact can be achieved through either hybrid \citep{dai2014whole, dai_optimizing_2012} or contact-implicit trajectory optimization \citep{ Mordatch12,patel_contact-implicit_2019,posa2014direct}. In the hybrid case, contact is modeled by specifying end-effector location at contact and defining constrained dynamics for each mode. Robustness to contact uncertainty has been studied by sampling contact locations and minimizing an expected cost \citep{dai_optimizing_2012, seyde_locomotion_2019}, by using Bayesian optimization to learn a robust cost function \citep{yeganegi_robust_2019}, and by constraining the risk of slipping \citep{shirai_risk-aware_2020}. However, developing general methods for contact uncertainty is difficult within the hybrid optimization framework as contact conditions are specified in the dynamical modes. 

In contrast, contact-implicit methods specify contact through a complementarity model which includes the nearest contact distance and friction coefficient \citep{Stewart96, posa2014direct}, and thus may provide a natural avenue for representing and planning for uncertainty in contact. Despite this potential, there have been few works exploring contact uncertainty within the contact-implicit framework. In \citep{mordatch2015ensemble}, contact point locations were sampled and an expected cost was minimized to produce robust motions. Recently, uncertainty in contact was modeled using probabilistic residual functions, and the expected residual was added to the cost to produce contact-sensitive trajectories \citep{drnach_robust_2021}, at the expense of producing potentially infeasible trajectories as uncertainty increased.   

\subsection{Chance Constraints}
To trade-off between robustness and constraint satisfaction, chance constraints can be added to an optimization problem to enforce a probabilistic version of the uncertain constraints \citep{celik_chance-constrained_2019, paulson_stochastic_2020, mesbah_stochastic_2016}. Chance constraints model uncertainty by defining a probability of constraint satisfaction, which can be tuned to enforce a conservative constraint or to relax the constraint. Previous works have  achieved robust vehicle trajectory planning under obstacle \citep{blackmore2011} and agent \citep{wang_non_gaussian} uncertainty using chance constraints. Chance constraints have also been applied to robot locomotion to increase the likelihood of avoiding collision with obstacles in uncertain locations \citep{gazar_stochastic_2020}, or to model slipping risk due to errors in the friction model \citep{shirai_risk-aware_2020, brandao_material_2016}. In contrast to collision avoidance, intermittent contact with the environment is required for robot locomotion, and while chance constraints have been applied to parts of the contact problem, they have yet to be applied to the full complementarity constraints for contact. Here, we investigate if chance constraints can trade-off between constraint satisfaction and robustness under contact uncertainty by combining them with our previously developed robust objectives \citep{drnach_robust_2021}.

\section{Problem Formulation}
 In this section, we present a robust contact-implicit trajectory optimization with both contact-robust costs and chance constraints to provide robustness to contact uncertainties while maintaining the feasibility of physical contact models.
\subsection{Contact-Implicit Trajectory Optimization} 
Planning robot motions that are subject to contact reaction forces can be achieved through contact-implicit trajectory optimization \citep{posa2014direct}. The traditional problem solves for generalized positions $q$, velocities $v$, controls $u$, and contact forces $\lambda$ through a discretized optimal control problem:
\begin{subequations} \label{contact-implicit}
    \begin{equation} \label{eq:OptimalControl}
        \min_{\mathbf{h, q, v, u, \lambda, \gamma}} \sum_{k=0}^{K-1} h_k L(x_k, u_k, \lambda_k)
    \end{equation}
    \begin{numcases} {\text{s.t.}} 
        \label{eq:BoundaryConstraint}
        x_0 = x(0), x_K = x(T_f) \\ \label{eq:Dynamics}
        M(v_{k+1} - v_{k}) + C = Bu_{k+1} + J_c^\top\lambda_{k+1} \\  \label{eq:DistanceConstraint}
        0 \leq \lambda_{N, k+1} \perp \phi(q_{k+1}) \geq 0 \\ \label{eq:SlidingConstraint} 
        0 \leq \lambda_{T,k+1} \perp \gamma_{k+1} + J_Tv_{k+1}  \geq 0 \\ \label{eq:FrictionConstraint}
        0 \leq \gamma_{k+1} \perp \mu\lambda_{N,k+1} - e^\top\lambda_{T,k+1} \geq 0 \\ \nonumber
        \hspace{1.5in} \forall k\in\{0,...,K-1\}
    \end{numcases}
\end{subequations}
where $L$ is the running cost, $h_k$ is the timestep, $x=(q,v)$ is the state, Eq.~\eqref{eq:BoundaryConstraint} are boundary constraints, $M$ is the generalized mass matrix, $C$ contains Coriolis and conservative force effects, $B$ is the control selection matrix, $J_c$ is the contact Jacobian, $\lambda_N$ and $\lambda_T$ are the normal and tangential contact reaction forces, $\phi$ is the contact distance, $\gamma$ is a slack variable corresponding to the magnitude of the sliding velocity, $\mu$ is the coefficient of friction, and $e$ is a matrix of 1s and 0s.

The contact Jacobian can be decomposed into normal and tangential components, $J_c^\top = [J_N^\top, J_T^\top]$. The normal component $J_N^\top$ maps the normal reaction force at the contact point to the generalized joint torques and is derived by projecting the contact point Jacobian onto the surface normal at the nearest contact point. The tangential component $J_T^\top$ maps the frictional forces at the contact point to generalized torques, and is the projection of the contact point Jacobian onto the plane tangent to the contact surface at the nearest point of contact. 

Equations \eqref{eq:DistanceConstraint}-\eqref{eq:FrictionConstraint} are complementarity constraints governing intermittent contact with the environment, where the notation $0\leq a \perp b \geq 0$ represents the complementarity constraints $a \geq 0, b\geq 0, ab = 0$. Equation \eqref{eq:DistanceConstraint} enforces that normal contact reaction forces are only imposed when the distance between the two objects is zero. Likewise, \eqref{eq:SlidingConstraint} and \eqref{eq:FrictionConstraint} govern the sticking and sliding phases of friction; when in sliding, \eqref{eq:FrictionConstraint} forces the friction forces to the edge of the friction cone and \eqref{eq:SlidingConstraint} requires $\gamma$ and the corresponding relative tangential velocities to be nonzero. In sticking, however, \eqref{eq:FrictionConstraint} forces the variable $\gamma$ to zero and \eqref{eq:SlidingConstraint} requires the corresponding relative tangential velocity to also be zero. We replaced the friction cone with a polyhedral approximation \citep{Stewart96}, denoted by the use of the $e$ in \eqref{eq:FrictionConstraint}, which contains only 1s and 0s,  instead of the use of the 2-norm, and we consider $\lambda_T$ to be the non-negative components of the friction force projected onto the polyhedron. The polyhedral approximation presented here can readily extend to the full three-dimensional case, although we do not study three-dimensional contact in this work. 

In general, the running cost is a function of all the decision variables, including the timesteps, states, controls, and reaction forces. However, in this work, we use a quadratic function of only the states and controls:
\begin{align*}
    L(x_k,u_k, \lambda_k) =& (x_k - x(T_f))^\top Q (x_k -x(T_f)) + u_k^\top R u_k.
\end{align*}
where R is the weight matrix on the control effort and Q is the weight matrix on the deviation from the final state. Our initial cost design does not depend on the reaction forces $\lambda$, although this is purely a design choice. Quadratic costs are common in the optimal control literature \citep{posa2014direct, Kuindersma16, patel_contact-implicit_2019}, although other cost functions can be used, such as the cost of transport \citep{posa2014direct}.

Problem \eqref{contact-implicit} is a mathematical program with equilibrium constraints, a type of nonlinear program (NLP) that can be difficult to solve. Two approaches to solve the problem numerically using standard NLP solvers like SNOPT \citep{GilMS05} include relaxing the complementarity constraints $ab \leq \epsilon$ (Figure \ref{fig:geometry_cc}D) and solving the problem from progressively smaller values of $\epsilon$ \citep{scholtes_convergence_2001, posa2014direct, manchester_contact-implicit_2019}, and replacing the constraints with an exact penalty term $\rho ab$ in the cost, where $\rho$ is chosen sufficiently large to drive the term $ab$ to zero \citep{baumrucker_mpec_2009, patel_contact-implicit_2019}. In this work, we found that the choice to use either the $\epsilon$-relaxation method or the exact penalty method was problem dependent. We also note that the robust cost we use is a probabilistic variant of the penalty method.

\subsection{Expected Residual Minimization}
The complementarity constraints in \eqref{contact-implicit} assume that perfect information about the contact model is available. However, if any of the model parameters are uncertain, the problem has stochastic complementarity constraints (SCP) \citep{luo_SCP_2013}  
$
    0 \leq z \perp F(z, \omega) \geq 0, \; \omega\in\Omega \label{eq:SCP}
$
where $\omega$ represents a random variable  on probability space $(\Omega, \mathcal{F}, \mathcal{P})$, $z$ is a deterministic variable, and $F(\cdot)$ is a vector-valued stochastic function. 

Prior works on SCPs
\citep{chen_robust_2009, tassastochastic, luo_SCP_2013} commonly replace the complementarity constraint with a residual function $\psi$ that attains its roots when the complementarity constraints are satisfied: 
$
    \psi(z, F) = 0 \Longleftrightarrow z \geq 0, F \geq 0, z F = 0
$. 
Although this formulation is for scalars $z$ and $F$, it generalizes to the case when $z$ and $F$ are vectors by applying the complementarity constraints and/or the residual function elementwise. In the Expected Residual Minimization (ERM) approach \citep{tassastochastic,chen_robust_2009}, the expected squared residual is minimized:
\begin{equation}\label{eq:ERM}
    \min_{z} \mathbb{E}[\norm{\psi(z, F(z, \omega))}^2]
\end{equation}
One advantage of the ERM is that its solutions have minimum sensitivity to random variations in the parameters \citep{chen_robust_2009}. 

Prior work using an ERM cost to plan for uncertainty in contact resulted in solutions that were robust to variations in the contact parameters \citep{drnach_robust_2021}. However, while the ERM method produced robust trajectories, as contact uncertainty increased, it also produced trajectories which were infeasible with respect to the expected values of the constraints. In this work, we use an ERM cost for Gaussian-distributed friction coefficient and normal distance \citep{tassastochastic, drnach_robust_2021}, and we add the ERM  to the running cost as:
\begin{equation}\label{eq:ERM_Cost}
    \min_{\mathbf{z = \{x,u,\lambda}\}} \sum_{k=0}^{K-1} \left(L(x_{k}, u_{k}, \lambda_{k}) + \alpha\mathbb{E}[\norm{\psi(z_k), F(z_{k}, \omega))}^2]\right)
\end{equation}
where $\alpha$ is a  penalty weighting factor selected to keep the ERM cost a few orders of magnitude higher than the other cost terms, as in the penalty method. In \eqref{eq:ERM_Cost}, the variable $z_k$ and the function $F$ are generic decision variables and constraint functions, respectively. In our work, we consider uncertainty in the terrain geometry and in the friction coefficient separately. In the case of uncertain terrain geometry, $F$ is the normal distance function $\phi(q)$ and $z$ includes the normal forces $\lambda_N$. Likewise, in the case of uncertainty in friction, $F$ is the linearized friction cone in \eqref{eq:FrictionConstraint} and $z$ includes the sliding velocity slack variable $\gamma$.

\subsection{Chance Complementarity Constraints} \label{linearGaussian}
\begin{figure}
    \centering
    \includegraphics[width =\linewidth]{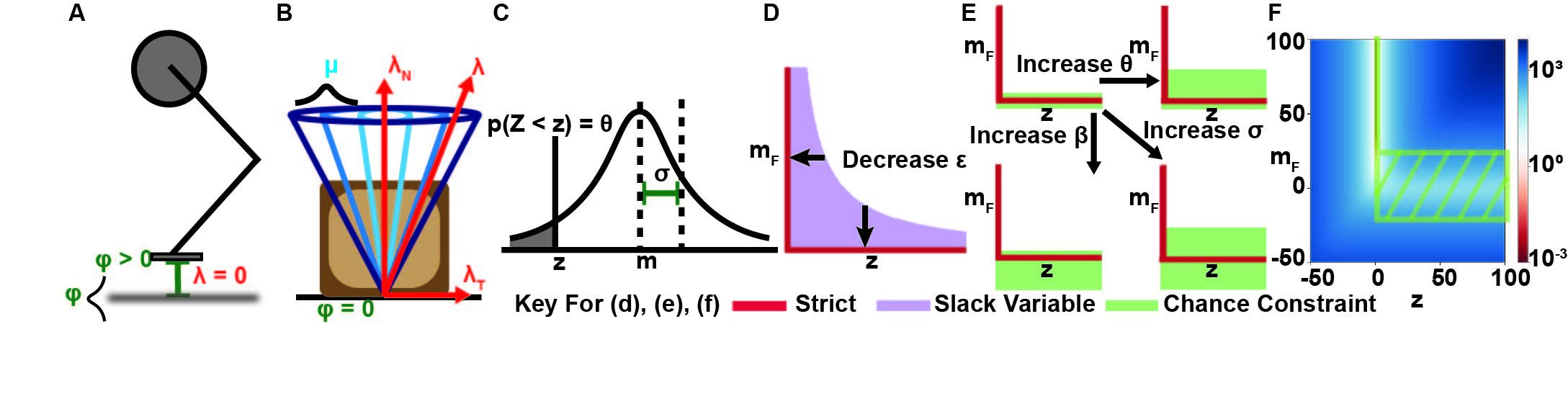}
    \caption{(a),(b) Contact geometry of the hopper and block examples, respectively, with uncertainty in (a) terrain height and (b) friction coefficient. (c) Gaussian distribution with mean $m$ and standard deviation $\sigma$, where $p(Z < z) = \theta$. (d) Relaxed complementarity constraint region for comparison with (e) chance complementarity constraint feasible regions for different risk bounds. (f) Overlap between ERM cost map and chance relaxed feasible region at $\sigma = 10$.  At high uncertainty, low ERM values approach the positive $m_F$ axis and the chance constraint region widens around the non-negative $z$ axis. }
    \label{fig:geometry_cc}
\end{figure}
Chance constraints are another general method for encoding uncertainty into constraints. Optimization with chance constraints enforces that the constraint is satisfied to within some user-specified probability, 
$
    \text{Pr}(z\in\mathcal{Z}) \geq 1 - \theta
$, where $\mathcal{Z}$ is the constraint set and $\theta$ is the specified probability of violation (Figure \ref{fig:geometry_cc}c). In this as in other works, we assume that $z$ is Gaussian, $z\sim\mathcal{N}(\mu_z, \Sigma)$, and that the constraint is linear, $\mathcal{Z} = \{z|c^\top z \leq b\}$ \citep{blackmore2011}. In this case, we can write the chance constraint using the error function \textbf{erf}\citep{celik_chance-constrained_2019}:
\begin{equation} \nonumber 
     \text{Pr}(c^Tz\leq b) = \frac{1}{2}\left(1 + \text{erf}\left(\frac{b - c^Tm_z}{\sqrt{2c^T\Sigma c}}\right)\right) \geq 1 - \theta  \\ \label{eq:GaussianChanceConstraint}
     \implies c^Tm_z \leq b - \sqrt{2c^T \Sigma c} \;\text{erf}^{-1}(1 - 2\theta)
\end{equation}
As $\textbf{erf}^{-1}$ takes values in $(-1,1)$, Eq. \eqref{eq:GaussianChanceConstraint} can represent either a relaxed ($\theta > 0.5$) or a conservative ($\theta < 0.5$) constraint.

To complement the robust ERM approach, in this work we investigate contact uncertainty by converting the stochastic complementarity constraints to deterministic, chance complementarity constraints. As with the Gaussian ERM, we assume the complementarity function is normally distributed $F \sim \mathcal{N}(m_F, \sigma^2)$, and we place probabilistic requirements on the components of the complementarity constraints $\text{Pr}(F \geq 0) \geq 1 - \beta$ and $\text{Pr}(zF \leq 0) \geq 1 - \theta$. Assuming that $z$ is a deterministic variable, by Eq. \eqref{eq:GaussianChanceConstraint} we have the following chance-complementarity constraints:
\begin{equation}\nonumber
     z \geq 0, \quad 
    m_F \geq -\sqrt{2}\sigma \; \text{erf}^{-1}(2\beta - 1), \\ \nonumber
    zm_F \leq - \sqrt{2}z\sigma\;\text{erf}^{-1}(1-2\theta)
\end{equation}

\begin{remark}
If either $\sigma=0$ or $\beta=\theta=0.5$, then the chance constraints recover the strict complementarity constraints.
 
\end{remark}
\begin{remark}
If $\beta = 0.5$ and $\theta > 0.5$, we recover a relaxed version of the complementarity constraints (Figure \ref{fig:geometry_cc}e):
$
    z \geq 0, \; m_F \geq 0, \; zm_F \leq \epsilon
$
where $\epsilon = -\sqrt{2}z\sigma\;\text{erf}^{-1}(1-2\theta) > 0$.
\end{remark}
\begin{remark}
If $\beta \geq 1 - \theta, z > 0$, the chance constraints relax the complementarity constraints into a tube around the mean:
\begin{align*}
    -\sqrt{2}\sigma\;\text{erf}^{-1}(2\beta - 1) \leq m_F \leq -\sqrt{2}\sigma \;\text{erf}^{-1}(1-2\theta)
\end{align*}
Note that, in this case, the chance constraints provide potentially asymmetric upper and lower bounds on the constraint violation, as by assumption $z>0$. For example, if $m_F$ and $z$ represent the normal distance and normal force, the chance constraints provide upper and lower bounds for the distance at which a non-zero normal force can be applied.
\end{remark} 

We also note that chance constraints \textit{cannot} provide robustness by making the complementarity constraints more conservative, as the original constraints have an empty interior. In contrast, previous works have used chance constraints to achieve robustness to uncertainty by removing part of the interior of the constraint set, making the constraint more conservative \citep{shirai_risk-aware_2020, gazar_stochastic_2020}. Chance complementarity constraints, however, always provide a relaxation of the original constraints, and give a probabilistic interpretation to previous methods using relaxed constraints \citep{manchester_contact-implicit_2019, patel_contact-implicit_2019}.

The chance complementarity constraints presented here possess nonempty solution sets only when $\beta > 1-\theta$; however, we note that not every choice of $\beta$ and $\theta$ is recommended, as choosing $\theta > 0.5$ and $\beta < 0.5$ requires the mean value $m_F$ to be strictly positive, whereas choosing $\theta < 0.5$ forces the mean $m_F$ to be strictly negative, both of which induce a bias into the complementarity problem. Therefore, we recommend further restricting the choice of parameter values to $\beta, \theta \geq 0.5$, as this choice ensures the mean $m_F$ can be zero, but still allows $m_F$ to take on positive and negative values.

In this work, we apply the chance constraints to relax the friction cone constraint (Eq. \eqref{eq:FrictionConstraint}) and the normal distance constraint (Eq. \eqref{eq:DistanceConstraint}), assuming normal distributions over the friction coefficient and the normal distance. We also include the corresponding ERM cost to examine the effects of chance constraints on the robustness of ERM solutions. We note that the failure probabilities $\beta, \theta$ can also be interpreted as \textit{risk bounds} \citep{shirai_risk-aware_2020}. By varying these risk bounds, we examine the tradeoff between strict feasibility under the expected value of the constraint when $\beta, \theta = 0.5$ and robustness to parameter variations under the ERM cost when $\beta, \theta > 0.5$.

\subsection{Quantifying Feasibility}
To quantify the feasibility of our solutions, we adopt a modified merit function $\mathcal{M}(z)$ \citep{seyde_locomotion_2019}:
\begin{equation} \label{eq:merit}
    \mathcal{M}(z) = \frac{1}{K}\sum_{k=0}^{K-1} \left(g_{EC, k}(z)^2 + \min(0, g_{IC, k}(z))^2\right)
\end{equation}
where $g_{EC}$ are the equality constraints, $g_{IC}$ are the inequality constraints, and $z$ are the decision variables. Here, the merit score only penalizes constraint violation, and provides a quantification of the \textit{feasibility} of the solutions. For the purposes of this study, we focus solely on contact feasibility under the expected value of the uncertain contact parameters, and apply the merit score to the friction cone constraint (Eq. \eqref{eq:FrictionConstraint}) for frictional uncertainty and to the normal distance constraint (Eq. \eqref{eq:DistanceConstraint}) for contact distance uncertainty. 

\section{Simulation Experiments}
We compared the chance-constrained risk-sensitive optimization approach to the ERM-only risk-sensitive approach \citep{drnach_robust_2021} and the traditional non-robust optimization approach in two experiments:
 a block sliding over a surface with uncertain friction and a single-legged hopper robot hopping over a flat terrain with uncertain height. All our examples were implemented in Python 3 using Drake \citep{drake} and solved using SNOPT \citep{GilMS05} to major optimality and feasibility tolerances of $10^{-6}$. Unless otherwise noted, all of our robust and chance-constrained problems were initialized with the reference, non-robust solution, and we used the same value for uncertainty $\sigma$ in the ERM objective as in the chance-constraints. Our code is available at \url{https://github.com/GTLIDAR/ChanceConstrainedRobustCITO}.

\subsection{Sliding Block with Uncertain Friction} Our first example is a planar 1m, 1kg cube sliding over a surface with nonzero friction (Figure \ref{fig:geometry_cc}B).  The state of the system $x = [p_{\rm CoM}, v_{\rm CoM}]$ includes the planar position and velocity of the center of mass of the block, $p_{\rm CoM}$ and $v_{\rm CoM}$ respectively, and the control $u$ is a horizontal force applied at the center of mass. We optimized for a 1s trajectory, discretized with 101 knot points, to travel between the initial state $x_0 = [0,0.5,0,0]^\top$ and final state $x_N = [5, 0.5, 0, 0]^\top$. The running cost had weight matrices $R = 10$ and $Q = \text{diag}([1,1,1,1])$. We first solved the optimization to a tolerance of $10^{-6}$ and then to $10^{-8}$; in this example, solving to the tighter tolerance improves the visual quality of the solutions. In the reference trajectory, we used friction coefficient $\mu=0.5$. For the uncertain cases, we assumed a mean friction of $\bar{\mu}=0.5$ and tested under 5 uncertainties $\sigma\in\{0.01, 0.05, 0.10, 0.30, 1.00\}$. When including chance constraints, we tested several combinations of the risk bounds $\theta, \beta \in \{0.51, 0.6, 0.7, 0.8, 0.9\}$ For completeness, we also tested the chance constraints without the ERM cost for uncertainties $\sigma \in \{0.1, 1.0\}$. We quantified the feasibility of our motion plans using the merit score (Eq. \eqref{eq:merit}) with the expected friction cone constraint (Eq. \eqref{eq:FrictionConstraint}), and we quantified the robustness using the maximum sliding velocity, as a higher velocity indicates less time in sliding. 

We evaluated the performance of the non-robust reference controls, the ERM controls, and the ERM with chance constraints controls in open-loop time-stepping simulations \citep{Stewart96}. To evaluate the robustness, we perturbed friction with 4 values uniformly spaced between $\mu = 0.3$ and $\mu = 0.7$ and evaluated the control performance as the difference between the block position at 1s and the target position. We quantified robustness as the range of final position errors under all friction perturbations.  We further evaluated the effect of the risk bounds on performance by first testing the chance constraints across a range of friction uncertainties with $\theta, \beta = 0.7$. We also evaluated the performance of the chance constraints at high uncertainty $(\sigma = 1.0)$  by testing 9 combinations of $\beta, \theta\in\{0.51, 0.7, 0.9\}$. 

\subsection{Single-Legged Hopper over an Uncertain Terrain}
Our second example is a 2D single-legged hopper with collision points at the toe and heel. The configuration $q$ includes the planar position (horizontal and vertical) of the base  $p_{\rm CoM}$ and the angles of the hip $\theta_H$, knee $\theta_K$, and ankle $\theta_A$; that is, $q=[p_{\rm CoM}, \theta_H, \theta_K, \theta_A]$. Thus, the state vector is $x = [q, \dot{q}]$, and the controls are the torques on the hip, knee, and ankle joints. In this example, the hopper travels $4$m in $3$s starting and ending at rest with the base 1.5m above the heel. We used 101 knot points and cost weights  $R = \text{diag}([0.01, 0.01, 0.01])$ and $Q = \text{diag}([1,10,10,100,100,1,1,1,1,1])$. 

We first solved for the reference trajectory using the exact penalty cost method to enforce the complementarity constraints for contact \citep{baumrucker_mpec_2009, patel_contact-implicit_2019}, and we initialized the reference optimization by linearly interpolating between the start and goal states. In our experiments with uncertainty, we assumed known friction coefficient $\mu = 0.5$ and uncertain terrain height with expected distance between initial hopper base height and terrain of $1.5$m. We tested the ERM and ERM with chance constraints approaches under 6 uncertainties roughly logarithmically spaced between $\sigma = 0.001$ and $\sigma = 0.5$ m. To more effectively utilize the ERM cost at high uncertainty, we scaled the normal distance by 10 during optimization, expressing the distance and its uncertainty in decimeters. At each uncertainty level, we tested 5 values of the chance parameters, $\theta \in \{0.51, 0.60, 0.70, 0.80, 0.90\}$, with $\beta=0.5$ in all cases to ensure no ground penetration. Note that when we apply chance constraints, we do not apply any other relaxation to the complementarity constraints. Instead, we use the strictly feasible solution from our progressive tightening procedure to warm-start the optimization with chance constraints. We quantified the feasibility of the hopping motion plans using the merit score (Eq. \eqref{eq:merit}) and the distance constraint (Eq. \eqref{eq:DistanceConstraint}). We used average foot height to quantify robustness, as higher foot heights indicate the hopper is less likely to trip over unexpected variations in ground height.

\section{Results}
\subsection{Chance Constraints Improve Friction Feasibility under High Uncertainty}
\begin{figure}
    \centering
    \includegraphics[width=\linewidth]{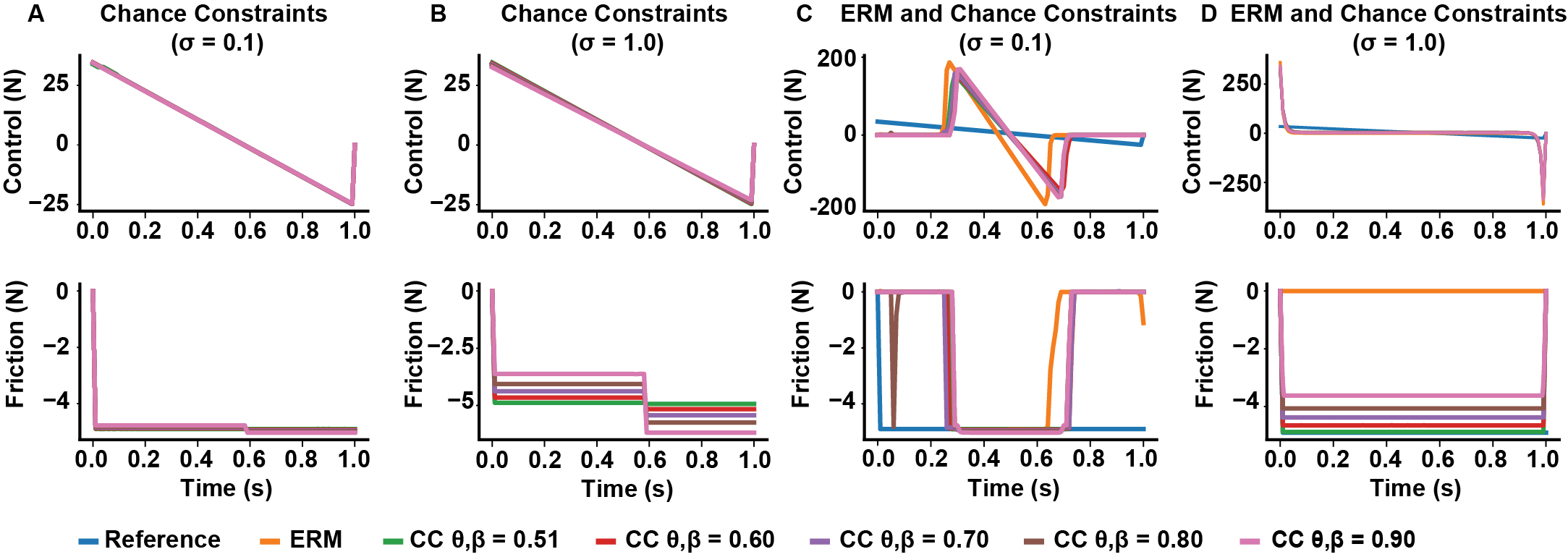}
    \caption{Effects of including chance constraints on contact-robust optimization at different uncertainty levels, for different risk bounds. (a, b) Including chance constraints without a robust cost, such as the ERM, does not have much effect on the optimized open-loop control, but can allow the friction force to vary under high uncertainty. (c, d) Including chance constraints with a contact robust cost has little effect on the robust solution at low uncertainty, but tightening the risk bounds $\theta$ and $\beta$ increases the friction force magnitude at high uncertainty.  
    }
    \label{fig:BlockTrajectories}
\end{figure}
In the sliding block example, optimizing under moderate uncertainty $(\sigma = 0.1)$ using chance constraints without the ERM cost produced trajectories that were nearly identical to the reference trajectory  (Figure \ref{fig:BlockTrajectories}A). When $\sigma  = 1.0$, however, the friction forces varied both above and below the reference value of -4.9N, demonstrating that chance constraints relax the friction cone around both sides of the mean. However, the optimized control was still nearly identical to the reference control (Figure \ref{fig:BlockTrajectories}B), indicating chance constraints alone may not offer any robustness to uncertainty in contact. 

In our optimizations combining the ERM with chance constraints, when the friction uncertainty was $\sigma < 0.1$, the ERM with chance constraints method produced friction forces around 4.9N during sliding, similar to those produced by the ERM method alone (Figure \ref{fig:BlockTrajectories}C). However, when the uncertainty was large ($\sigma = 1.0$), the ERM produced friction forces at 0N during the entire motion, which is infeasible for all friction coefficients except $\mu=0$. In contrast, the ERM with chance constraints produced nonzero friction forces, and the magnitude of the friction forces increased as the risk bounds decreased and converged towards the expected value for friction at 4.9N (Figure \ref{fig:BlockTrajectories}D), indicating a solution with improved feasibility under the expected friction coefficient.

\begin{figure}
    \centering
    \includegraphics[width=\linewidth]{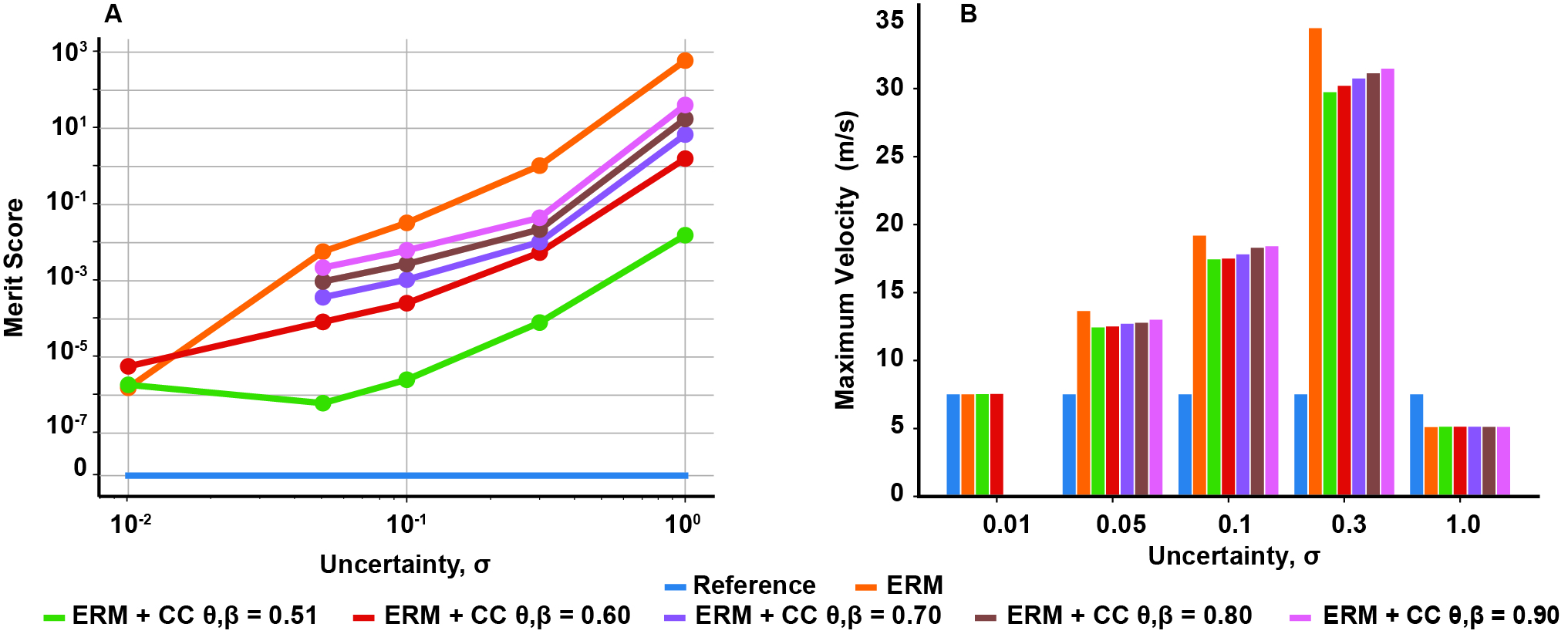}
    \caption{Chance constraint mediated trade-off between expected friction cone feasibility and robustness to friction uncertainty (signified by maximum sliding velocity). (a) Merit scores across uncertainty and risk tolerances, quantifying violation of the expected friction cone constraint. (b) Maximum sliding velocity across uncertainty and risk tolerances, signifying robustness as larger velocities indicate shorter sliding times. Both constraint violation and maximum velocity increase with increasing uncertainty and with increasing risk bounds. Missing data points indicate the optimization was not solved successfully.}
    \label{fig:BlockMerit}
\end{figure}

Across all uncertainties, the solutions of the ERM and ERM with chance constraints tended to improve in friction cone feasibility as the uncertainty decreased, as indicated by a decrease in the merit score (Figure \ref{fig:BlockMerit}A). Moreover, at any fixed uncertainty, the friction merit score decreased as the risk parameters decreased, with the ERM-only solution and reference solution acting as upper and lower bounds, respectively. Similarly, the maximum sliding velocity of the block increased with increasing uncertainty, indicating less sliding time under uncertainty, but decreased with decreasing the risk parameters (Figure \ref{fig:BlockMerit}B), except in the highest uncertainty case. The range of maximum velocity across chance parameters also increased with increasing uncertainty, from 0.02m/s at $\sigma = 0.01$ to 1.73m/s at $\sigma = 0.3$. However, at the highest uncertainty, the sliding velocity for the ERM and chance constraints were all identical and less than that of the reference. In the $\sigma = 1$ case, the ERM failed to provide robustness to friction uncertainty; in this case, the ERM does not model the friction cone constraint well, and allows the optimization to set the friction forces to zero. Without friction, the optimal control is an impulsive, bang-bang controller (Figure \ref{fig:BlockTrajectories}D) and the resulting trajectory has almost constant velocity at 5m/s. However, the addition of chance constraints did improve the feasibility of the final motion plans with respect to the friction cone constraint, but did not alter the sliding velocity. Taken together, these results indicate that the chance constraints can mediate a trade-off between the robustness to friction uncertainty provided by the ERM and the strict feasibility provided by the reference solution. 

\subsection{Chance constraints improve average performance against friction perturbations in simulation}
\begin{figure}
    \centering
    \includegraphics[width = 5.3in]{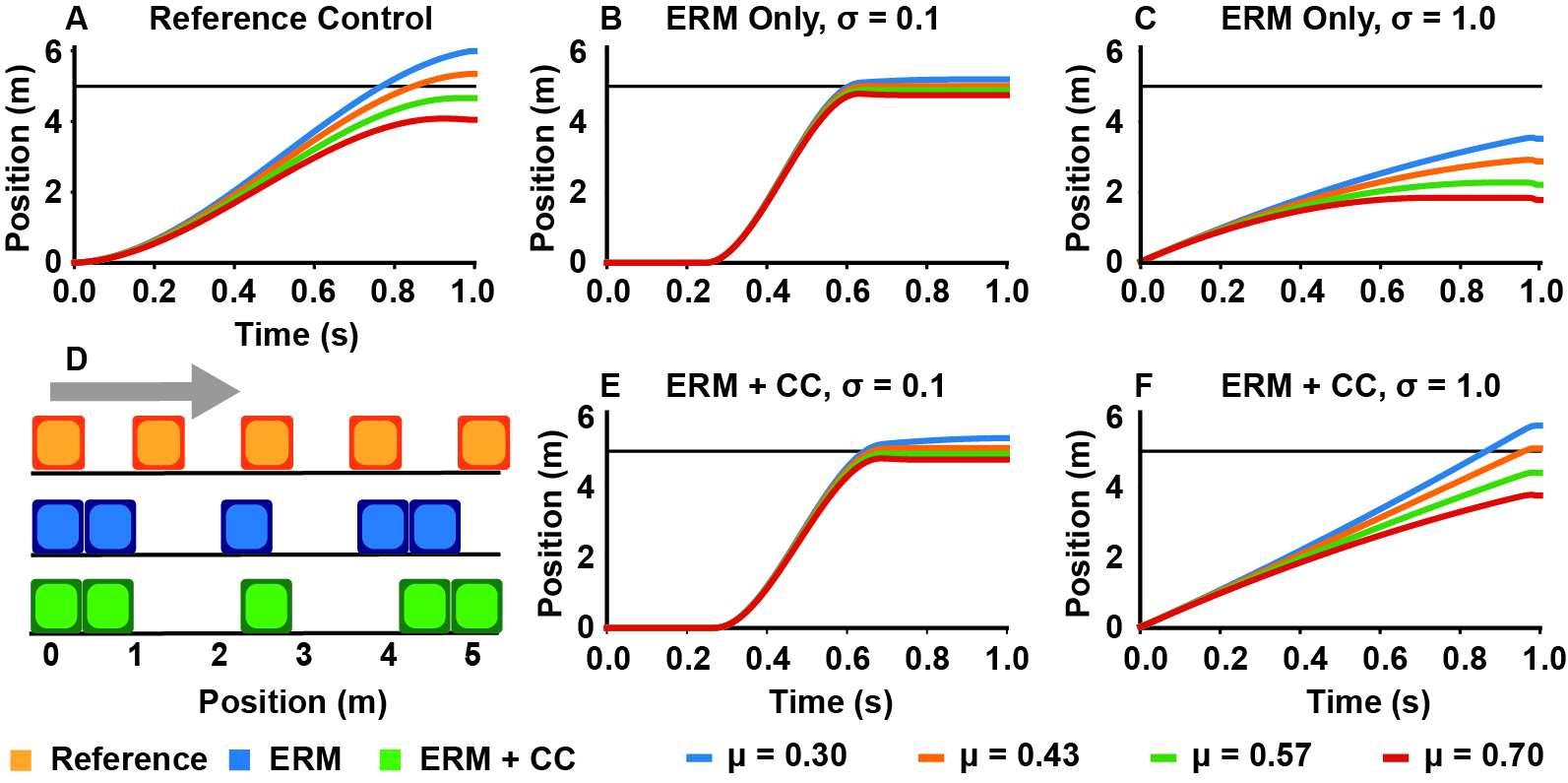}
    \caption{Example block simulations demonstrating chance constraints retain robustness at moderate uncertainty and improve feasibility performance at high uncertainty, compared to the (a) simulations using the reference controls, for four different values of the friction coefficient. Simulations using controls generated under only the contact-robust ERM cost result in a low spread around the desired position for moderate uncertainty (b), but can result in a large average position error when the friction uncertainty is large (c). Simulations using  controls generated using ERM with chance constraints maintain a low spread at moderate uncertainty (e), and have a low final position error at high uncertainty (f). (d) Illustration of the motion of the block for the reference, ERM, and ERM with chance constraint controls under high friction uncertainty.}
    \label{fig:BlockSimulation}
\end{figure}
\begin{figure}
    \centering
    \includegraphics[width=\linewidth]{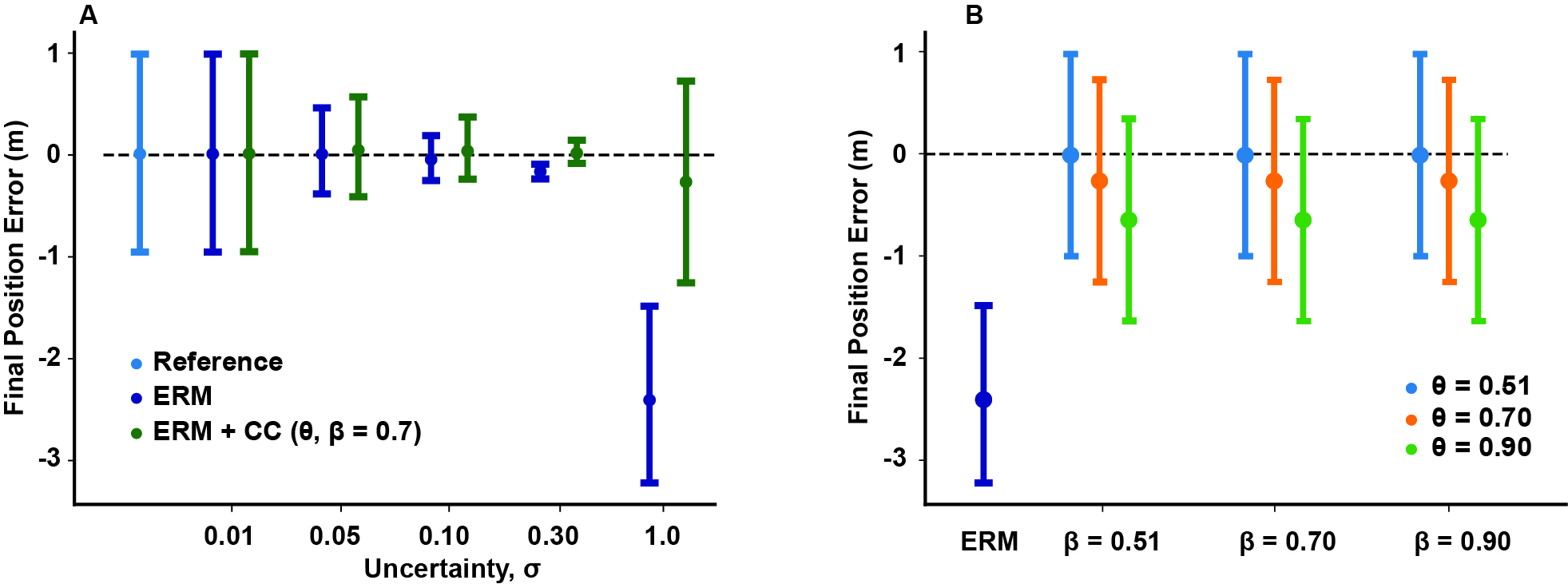}
    \caption{Effects of chance constraints on robustness of sliding block controls in open loop simulations. (a) Mean and range of final position errors for the ERM with and without chance constraints planned under different uncertainties, compared to those of the reference. The addition of chance constraints maintains the low range of final position errors produced by the ERM, but at high uncertainty the chance constraints reduce the average final position error. (b) Mean and range of final position error of simulated chance constraint controls under different risk tolerances compared to the mean and range for the ERM under the highest friction uncertainty case ($\sigma = 1.0)$. Increasing the upper risk bound $\beta$ has little effect, while increasing the lower risk bound
    $\theta$ can increase the average final position error.}
    \label{fig:BlockErrorPlot}
\end{figure}
In our open loop simulations with the block example, the controls generated under ERM with chance constraints performed similarly to those generated under only the ERM for uncertainties kess than $0.1$ (mean position error 0.04 and error range 0.44 for ERM only, mean -0.03 and range 0.61 for ERM with chance constraints at $\sigma =0.1$) (Figure \ref{fig:BlockSimulation}). However, at high uncertainty $\sigma = 1.0$, the ERM with chance constraint simulation achieved a lower average position error compared to the ERM alone (0.26 for chance constraints, 2.41 for ERM only), although both had a similar range of position errors (Figure \ref{fig:BlockErrorPlot}A). By varying the chance parameters during optimization, we found that changing $\beta$ had little effect on simulation results, while increasing $\theta$ resulted in a slight increase in the final position error, from an average error of 0.01 at $\theta = 0.51$ to 0.65 at $\theta=0.9$, for all values of $\beta$ (Figure \ref{fig:BlockErrorPlot}B). Moreover, changing $\theta$ and $\beta$ at high uncertainty had no effect on the range of final positions achieved, indicating again that the chance constraints modulate the feasibility of the motion plan, while the robustness is provided by the ERM cost.

\subsection{Chance constraints mediate the distance at which contact forces are applied}
\begin{figure}
    \centering
    \includegraphics[width=\linewidth]{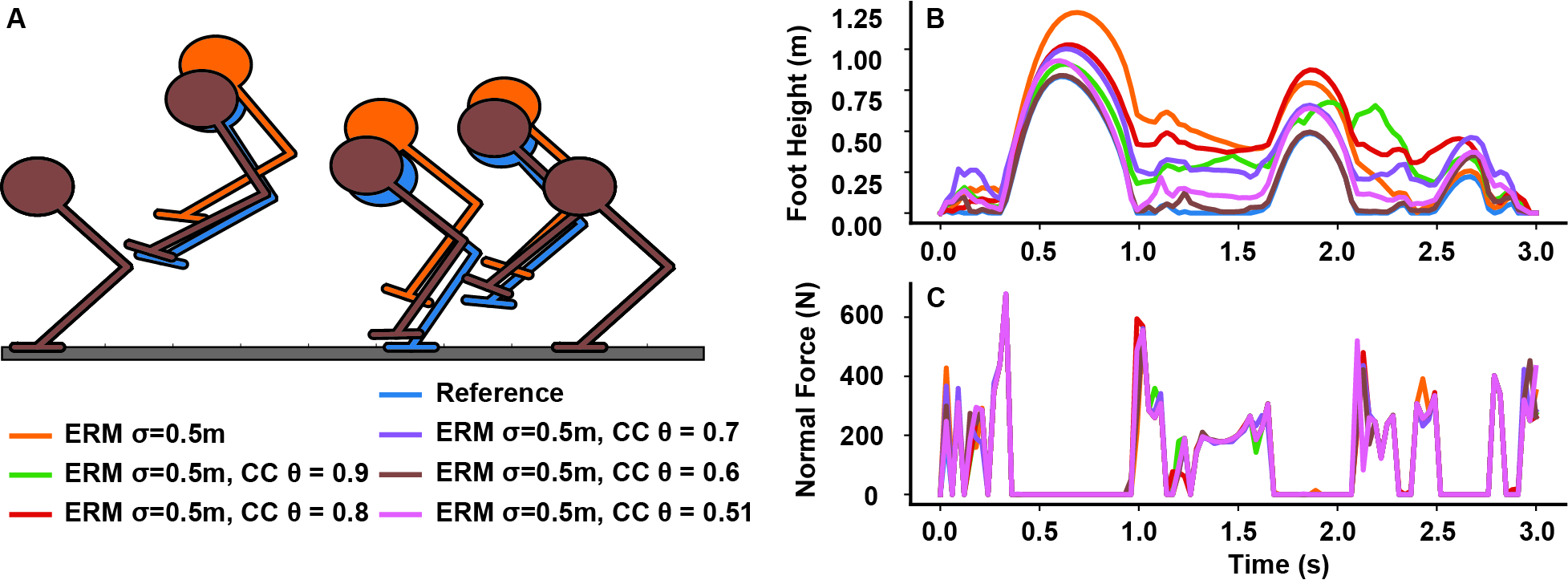}
    \caption{Effect of including chance constraints on hopping under distance uncertainty. (a) Selected frames of the hopper trajectory comparing the reference, non-robust trajectory, the ERM only trajectory, and the ERM with chance constraints trajectory. Only the $\theta = 0.6$ case is illustrated for brevity.  (b) Planned foot heights for the hopper under high distance uncertainty ($\sigma = 0.5$ m) for different risk bounds, compared to the ERM and reference trajectories, and (c) the associated normal ground reaction forces. The ERM cost allows for contact forces to be applied at nonzero distances; however, as the risk bounds decrease, the distance at which forces are applied also decreases.}
    \label{fig:hoppertrajectories}
\end{figure}
\begin{figure}
    \centering
    \includegraphics[width=\linewidth]{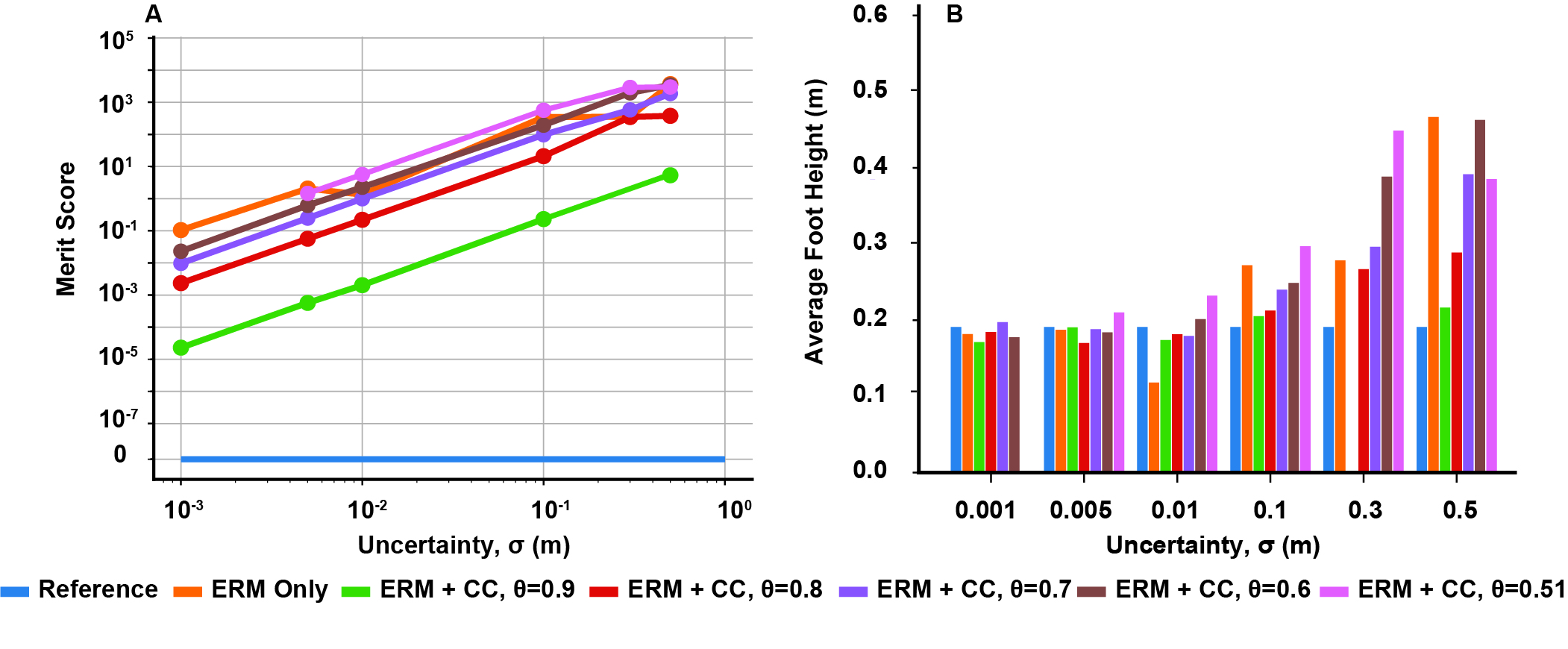}
    \caption{Chance constraint mediated trade-off between contact distance feasibility and average foot height for robustness. (a) Merit scores across distance uncertainty and risk bounds, quantifying the violation of the expected contact distance constraint. (b) Average foot height across uncertainty and risk bounds, where higher average height indicates more contact-robust hopping. Both constraint violation and maximum foot height increase with increasing uncertainty and with increasing risk bounds. Missing data points indicate the optimization was not solved successfully.}
    \label{fig:hoppermerit}
\end{figure}
In the hopping example with contact distance uncertainty, the ERM alone produced higher average foot height with increasing uncertainty, up to an average of 0.46m at our highest value of uncertainty ($\sigma = 0.5$ m). Introducing chance constraints, however, reduced the foot height and reduced the distance at which the contact normal forces were nonzero, and the decrease in foot height trended with decreasing the risk parameters $\theta, \beta$ (Figure \ref{fig:hoppertrajectories}B,C). Across all uncertainties and risk parameters, the chance constraints tended to reduce foot height as the risk parameters decreased, and the range of foot heights generated by the risk parameters tended to increase with increasing uncertainty (Figure \ref{fig:hoppermerit}B), although there are exceptions which could be due to the highly nonlinear and nonconvex nature of the problem. However, we note that neither the ERM nor the chance constraints had much effect on the optimized reaction forces; in this example, the effects were limited mainly to the contact distance. By using the merit score, we also observed that the contact distance infeasibility decreased with both decreasing uncertainty and decreasing the risk parameters (Figure \ref{fig:hoppermerit}A). While the reference case provides a lower bound for the infeasibility, as it did in the block example, in this example the ERM only trajectory was not strictly the upper bound for all uncertainties, although this may be due to the presence of multiple local minima in the optimization. 

\section{Discussion and Conclusions}
In this work we proposed a novel framework for accounting for contact uncertainty in trajectory optimization. As previously explored, the ERM cost represents a robust contact-averse objective but also results in infeasible trajectories as the contact uncertainty grows \citep{drnach_robust_2021}. Here we developed chance complementarity constraints to convert the stochastic constraints into deterministic constraints and showed that the chance constraints can mediate a trade-off between feasibility and robustness by changing the risk bounds $\theta$ and $\beta$. The improved feasibility is achieved because the chance constraints limit the region of allowable solutions to the ERM to those near the non-negative $m_F$ and $z$ axes, i.e. the solution set of the non-stochastic complementarity constraints; moreover, as the risk bounds are decreased, the allowable set approaches the complementarity solution under the mean values of the constraints, representing the limit of perfect feasibility under the mean but no robustness. 

Our work with chance-constraints is similar to previous works which have applied chance-constraints  to obstacle avoidance \citep{gazar_stochastic_2020} or to modeling frictional uncertainty \citep{shirai_risk-aware_2020} for locomotion. These works claim that the chance constraints provide a measure of robustness by using risk bounds to make the constraints more conservative, which can be thought of as making an obstacle larger or by making the friction cone narrower. This type of robustness is similar to worst-case robustness; the generated plan accounts for the worst possible constraint violations, but may still be sensitive to variations in the constraint parameters \citep{drnach_robust_2021}. In this work, we applied chance constraints to problems which require intermittent contact, and we noted that the complementarity constraints cannot be made more conservative as their solution sets have an empty interior. Instead, we demonstrated that chance constraints relaxed the contact constraints and improved the physical feasibility of trajectories generated with a robust cost; lower risk bounds produced trajectories which were feasible under the expected constraints but were potentially sensitive to variations, while higher risk bounds allowed trajectories to violate the expected constraints to achieve robustness. 

Here we considered solely the problem of accounting for uncertainty in contact during motion planning; we specifically have not investigated handling uncertainty in contact with control. Future work could convert our technique into a feedback control policy by re-planning in a receding horizon fashion; however, current methods for solving contact-implicit problems are too slow to be used reactively in real-time. Thus, advancements in efficient solvers for contact-implicit problems are necessary before our work can be used in a receding horizon control fashion, such as those used in hybrid optimization to generate gait libraries \citep{hereid2019rapid}. Apart from replanning, other  methods for controlling through contact have already been developed, including contact mode-invariant stabilizing control using Lyapunov analysis \citep{posa_stability_2016} and a risk-sensitive impedance optimization for handing control through uncertain contact \citep{hammoud_impedance_2021}. Although these approaches show promise for stabilizing and controlling locomotion through contact, the former has yet to be demonstrated on terrain with variations and the latter requires a reference trajectory with a contact schedule. The overarching goal of our work is to complement these approaches by generating a reference trajectory, including the contact sequence, which is robust to terrain variations. By planning trajectories which are robust to contact uncertainty - for example, by avoiding areas of the terrain with large variations - we aim to alleviate some of the burden on the controller and improve the overall performance of the system.

In this work, we parameterized uncertainty in the distance to the terrain and in the friction coefficient using Gaussian distributions, as this distribution provides analytical formulas for the ERM cost and for the chance constraints. Having access to analytical formulas means we only needed to generate one robust trajectory, instead of generating multiple samples to achieve robustness \citep{mordatch2015ensemble, seyde_locomotion_2019}. Given that generating a single trajectory using the contact-implicit approach requires substantial computation time, the analytical formulas saved us considerable computation time by avoiding solving the problem for multiple samples of the terrain geometry or friction coefficient. However, using the Gaussian distribution has distinct disadvantages in theory, as it places non-zero probability mass over regions which are physically impossible, such as over negative friction coefficients or over terrain heights which result in interpenetration (e.g. terrain heights that are above the current contact point location). Such physically impossible regions could be avoided in future works by using distributions over a subset of the reals, such as the truncated Gaussian distribution or the Gamma distribution. However, using such distributions might require considerable effort to evaluate the ERM cost and chance constraints, which have so far been developed largely for Gaussian distributed variables.

The main advantage of our chance-constrained ERM approach is that we can generate trajectories with varying degrees of robustness to contact uncertainty without changing the uncertainty. Thus, when faced with uncertain terrain, we can choose between being robust to terrain variations or being optimal with respect to our original objective without artificially changing the uncertainty in the model. Our work here focused on investigating these behaviors in simple systems on 2-dimensional terrain. In future works we could scale up our approach to full-scale robots traversing 3-dimensional terrain. We expect the complexity of solving the ERM and chance constraints to scale only with the number of contacts and not with the state dimension of the robot, as the number of complementarity constraints, and therefore the number of ERM costs and chance constraints, is linear in the number of contact points and not dependent on the state dimension - for example, adding several contact points to the sliding block and putting obstacles in the environment would make the contact problem more challenging, even though the state dimension is the same. Once we have scaled up to three dimensions, we could also evaluate our methods experimentally on full-scale robots, such as a quadruped, and compare the performance of our robust motion plans against the traditional approach using a simple controller, and against other risk-sensitive control approaches such as \citep{hammoud_impedance_2021}.

\bibliographystyle{unsrtnat}
\bibliography{references}  

\begin{thebibliography}{36}
\providecommand{\natexlab}[1]{#1}
\providecommand{\url}[1]{\texttt{#1}}
\expandafter\ifx\csname urlstyle\endcsname\relax
  \providecommand{\doi}[1]{doi: #1}\else
  \providecommand{\doi}{doi: \begingroup \urlstyle{rm}\Url}\fi

\bibitem[Dai et~al.(2014)Dai, Valenzuela, and Tedrake]{dai2014whole}
Hongkai Dai, Andr{\'e}s Valenzuela, and Russ Tedrake.
\newblock Whole-body motion planning with centroidal dynamics and full
  kinematics.
\newblock In \emph{IEEE-RAS International Conference on Humanoid Robots}, pages
  295--302, 2014.

\bibitem[Mordatch et~al.(2012)Mordatch, Todorov, and Popovi{\'c}]{Mordatch12}
Igor Mordatch, Emanuel Todorov, and Zoran Popovi{\'c}.
\newblock Discovery of complex behaviors through contact-invariant
  optimization.
\newblock \emph{ACM Transactions on Graphics (TOG)}, 31\penalty0 (4):\penalty0
  43--43:8, August 2012.

\bibitem[Winkler et~al.(2018)Winkler, Bellicoso, Hutter, and
  Buchli]{winkler_gait_2018}
Alexander~W. Winkler, C.~Dario Bellicoso, Marco Hutter, and Jonas Buchli.
\newblock Gait and {Trajectory} {Optimization} for {Legged} {Systems} {Through}
  {Phase}-{Based} {End}-{Effector} {Parameterization}.
\newblock \emph{IEEE Robotics and Automation Letters}, 3\penalty0 (3):\penalty0
  1560--1567, July 2018.
\newblock ISSN 2377-3766, 2377-3774.

\bibitem[Patel et~al.(2019)Patel, Shield, Kazi, Johnson, and
  Biegler]{patel_contact-implicit_2019}
Amir Patel, Stacey Shield, Saif Kazi, Aaron~M. Johnson, and Lorenz~T. Biegler.
\newblock Contact-implicit trajectory optimization using orthogonal
  collocation.
\newblock \emph{{IEEE} Robotics and Automation Letters}, 4\penalty0
  (2):\penalty0 2242--2249, 2019.

\bibitem[Toussaint et~al.(2014)Toussaint, Ratliff, Bohg, Righetti, Englert, and
  Schaal]{toussaint_dual_2014}
M.~Toussaint, N.~Ratliff, J.~Bohg, L.~Righetti, P.~Englert, and S.~Schaal.
\newblock Dual execution of optimized contact interaction trajectories.
\newblock In \emph{2014 {IEEE}/{RSJ} {International} {Conference} on
  {Intelligent} {Robots} and {Systems}}, pages 47--54, September 2014.
\newblock \doi{10.1109/IROS.2014.6942539}.

\bibitem[Gazar et~al.(2020)Gazar, Khadiv, Del~Prete, and
  Righetti]{gazar_stochastic_2020}
Ahmad Gazar, Majid Khadiv, Andrea Del~Prete, and Ludovic Righetti.
\newblock Stochastic and {Robust} {MPC} for {Bipedal} {Locomotion}: {A}
  {Comparative} {Study} on {Robustness} and {Performance}.
\newblock \emph{arXiv:2005.07555 [cs, eess]}, May 2020.

\bibitem[Posa et~al.(2014)Posa, Cantu, and Tedrake]{posa2014direct}
Michael Posa, Cecilia Cantu, and Russ Tedrake.
\newblock A direct method for trajectory optimization of rigid bodies through
  contact.
\newblock \emph{The International Journal of Robotics Research}, 33\penalty0
  (1):\penalty0 69--81, 2014.

\bibitem[Dai and Tedrake(2012)]{dai_optimizing_2012}
Hongkai Dai and Russ Tedrake.
\newblock Optimizing robust limit cycles for legged locomotion on unknown
  terrain.
\newblock In \emph{2012 {IEEE} 51st {IEEE} {Conference} on {Decision} and
  {Control} ({CDC})}, pages 1207--1213. IEEE, December 2012.

\bibitem[Carius et~al.(2018)Carius, Ranftl, Koltun, and
  Hutter]{carius_trajectory_2018}
Jan Carius, René Ranftl, Vladlen Koltun, and Marco Hutter.
\newblock Trajectory {Optimization} {With} {Implicit} {Hard} {Contacts}.
\newblock \emph{IEEE Robotics and Automation Letters}, 3\penalty0 (4):\penalty0
  3316--3323, October 2018.

\bibitem[Kuindersma et~al.(2016)Kuindersma, Deits, Fallon, Valenzuela, Dai,
  Permenter, Koolen, Marion, and Tedrake]{Kuindersma16}
Scott Kuindersma, Robin Deits, Maurice Fallon, Andr{\'e}s Valenzuela, Hongkai
  Dai, Frank Permenter, Twan Koolen, Pat Marion, and Russ Tedrake.
\newblock Optimization-based locomotion planning, estimation, and control
  design for {{Atlas}}.
\newblock \emph{Autonomous Robots}, 40\penalty0 (3):\penalty0 429--455, 2016.

\bibitem[Mordatch et~al.(2015)Mordatch, Lowrey, and
  Todorov]{mordatch2015ensemble}
Igor Mordatch, Kendall Lowrey, and Emanuel Todorov.
\newblock Ensemble-cio: Full-body dynamic motion planning that transfers to
  physical humanoids.
\newblock In \emph{IEEE/RSJ International Conference on Intelligent Robots and
  Systems}, pages 5307--5314, 2015.

\bibitem[Yeganegi et~al.(2019)Yeganegi, Khadiv, Moosavian, Zhu, Prete, and
  Righetti]{yeganegi_robust_2019}
M.~H. Yeganegi, M.~Khadiv, S.~A.~A. Moosavian, J.~Zhu, A.~Del Prete, and
  L.~Righetti.
\newblock Robust {Humanoid} {Locomotion} {Using} {Trajectory} {Optimization}
  and {Sample}-{Efficient} {Learning}*.
\newblock In \emph{{IEEE} {International} {Conference} on {Humanoid} {Robots}},
  pages 170--177, October 2019.

\bibitem[Ponton et~al.(2018)Ponton, Schaal, and Righetti]{ponton_role_2018}
Brahayam Ponton, Stefan Schaal, and Ludovic Righetti.
\newblock The {Role} of {Measurement} {Uncertainty} in {Optimal} {Control} for
  {Contact} {Interactions}.
\newblock \emph{arXiv:1605.04344 [eess]}, January 2018.
\newblock URL \url{http://arxiv.org/abs/1605.04344}.

\bibitem[Farshidian and Buchli(2015)]{farshidian_risk_2015}
Farbod Farshidian and Jonas Buchli.
\newblock Risk {Sensitive}, {Nonlinear} {Optimal} {Control}: {Iterative}
  {Linear} {Exponential}-{Quadratic} {Optimal} {Control} with {Gaussian}
  {Noise}.
\newblock \emph{arXiv:1512.07173 [cs]}, December 2015.

\bibitem[Hackett et~al.(2020)Hackett, Gao, Daley, Clark, and
  Hubicki]{hackett_risk-constrained_2020}
J.~Hackett, W.~Gao, M.~Daley, J.~Clark, and C.~Hubicki.
\newblock Risk-constrained {Motion} {Planning} for {Robot} {Locomotion}:
  {Formulation} and {Running} {Robot} {Demonstration}.
\newblock In \emph{{IEEE}/{RSJ} {International} {Conference} on {Intelligent}
  {Robots} and {Systems}}, pages 3633--3640, October 2020.

\bibitem[Shirai et~al.(2020)Shirai, Lin, Tanaka, Mehta, and
  Hong]{shirai_risk-aware_2020}
Yuki Shirai, Xuan Lin, Yusuke Tanaka, Ankur Mehta, and Dennis Hong.
\newblock Risk-{Aware} {Motion} {Planning} for a {Limbed} {Robot} with
  {Stochastic} {Gripping} {Forces} {Using} {Nonlinear} {Programming}.
\newblock \emph{IEEE Robotics and Automation Letters}, 5\penalty0 (4):\penalty0
  4994--5001, October 2020.
\newblock ISSN 2377-3766.

\bibitem[Drnach and Zhao(2021)]{drnach_robust_2021}
L.~Drnach and Y.~Zhao.
\newblock Robust {Trajectory} {Optimization} {Over} {Uncertain} {Terrain}
  {With} {Stochastic} {Complementarity}.
\newblock \emph{IEEE Robotics and Automation Letters}, 6\penalty0 (2):\penalty0
  1168--1175, April 2021.

\bibitem[Seyde et~al.(2019)Seyde, Carius, Grandia, Farshidian, and
  Hutter]{seyde_locomotion_2019}
Tim Seyde, Jan Carius, Ruben Grandia, Farbod Farshidian, and Marco Hutter.
\newblock Locomotion {Planning} through a {Hybrid} {Bayesian} {Trajectory}
  {Optimization}.
\newblock In \emph{2019 {International} {Conference} on {Robotics} and
  {Automation} ({ICRA})}, pages 5544--5550, May 2019.

\bibitem[Stewart and Trinkle(1996)]{Stewart96}
David~E Stewart and J~C Trinkle.
\newblock An implicit time-stepping scheme for rigid body dynamics with
  inelastic collisions and coulomb friction.
\newblock \emph{International Journal for Numerical Methods in Engineering},
  39\penalty0 (15):\penalty0 2673--2691, 1996.

\bibitem[Celik et~al.(2019)Celik, Abdulsamad, and
  Peters]{celik_chance-constrained_2019}
Onur Celik, Hany Abdulsamad, and Jan Peters.
\newblock Chance-{Constrained} {Trajectory} {Optimization} for {Non}-linear
  {Systems} with {Unknown} {Stochastic} {Dynamics}.
\newblock \emph{arXiv:1906.11003 [cs, eess]}, July 2019.

\bibitem[Paulson et~al.(2020)Paulson, Buehler, Braatz, and
  Mesbah]{paulson_stochastic_2020}
Joel~A. Paulson, Edward~A. Buehler, Richard~D. Braatz, and Ali Mesbah.
\newblock Stochastic model predictive control with joint chance constraints.
\newblock \emph{International Journal of Control}, 93\penalty0 (1):\penalty0
  126--139, January 2020.

\bibitem[Mesbah(2016)]{mesbah_stochastic_2016}
Ali Mesbah.
\newblock Stochastic {Model} {Predictive} {Control}: {An} {Overview} and
  {Perspectives} for {Future} {Research}.
\newblock \emph{IEEE Control Systems Magazine}, 36\penalty0 (6):\penalty0
  30--44, December 2016.
\newblock ISSN 1941-000X.

\bibitem[Blackmore et~al.(2011)Blackmore, Ono, and WIlliams]{blackmore2011}
Lars Blackmore, Masahiro Ono, and Brian~C. WIlliams.
\newblock Chance-constrained optimal path planning with obstacles.
\newblock \emph{IEEE Transactions on Robotics}, 27\penalty0 (6), 2011.

\bibitem[Wang et~al.(2020)Wang, Jasour, and Williams]{wang_non_gaussian}
Allen Wang, AshKan Jasour, and Brain~C. Williams.
\newblock Non-gaussian chance-constrained trajectory planning for autonomous
  vehicles under agent uncertainty.
\newblock \emph{IEEE Robotics and Automation Letters}, July 2020.

\bibitem[Brandão et~al.(2016)Brandão, Shiguematsu, Hashimoto, and
  Takanishi]{brandao_material_2016}
Martim Brandão, Yukitoshi~Minami Shiguematsu, Kenji Hashimoto, and Atsuo
  Takanishi.
\newblock Material recognition {CNNs} and hierarchical planning for biped robot
  locomotion on slippery terrain.
\newblock In \emph{2016 {IEEE}-{RAS} 16th {International} {Conference} on
  {Humanoid} {Robots} ({Humanoids})}, pages 81--88, November 2016.
\newblock ISSN: 2164-0580.

\bibitem[Gill et~al.(2005)Gill, Murray, and Saunders]{GilMS05}
Philip~E. Gill, Walter Murray, and Michael~A. Saunders.
\newblock {SNOPT}: An {SQP} algorithm for large-scale constrained optimization.
\newblock \emph{SIAM Rev.}, pages 99--131, 2005.

\bibitem[Scholtes(2001)]{scholtes_convergence_2001}
Stefan Scholtes.
\newblock Convergence {Properties} of a {Regularization} {Scheme} for
  {Mathematical} {Programs} with {Complementarity} {Constraints}.
\newblock \emph{SIAM Journal on Optimization}, 11\penalty0 (4):\penalty0
  918--936, January 2001.

\bibitem[Manchester et~al.(2019)Manchester, Doshi, Wood, and
  Kuindersma]{manchester_contact-implicit_2019}
Zachary Manchester, Neel Doshi, Robert~J Wood, and Scott Kuindersma.
\newblock Contact-implicit trajectory optimization using variational
  integrators.
\newblock \emph{The International Journal of Robotics Research}, 38\penalty0
  (12-13):\penalty0 1463--1476, October 2019.

\bibitem[Baumrucker and Biegler(2009)]{baumrucker_mpec_2009}
B.~T. Baumrucker and L.~T. Biegler.
\newblock {MPEC} strategies for optimization of a class of hybrid dynamic
  systems.
\newblock \emph{Journal of Process Control}, 19\penalty0 (8):\penalty0
  1248--1256, September 2009.
\newblock ISSN 0959-1524.

\bibitem[Luo and Lu(2013)]{luo_SCP_2013}
Mei-Ju Luo and Yuan Lu.
\newblock Properties of expected residual minimization model for a class of
  stochastic complementarity problems.
\newblock \emph{Journal of Applied Mathematics}, 2013:\penalty0 1--7, 2013.

\bibitem[Chen et~al.(2009)Chen, Zhang, and Fukushima]{chen_robust_2009}
Xiaojun Chen, Chao Zhang, and Masao Fukushima.
\newblock Robust solution of monotone stochastic linear complementarity
  problems.
\newblock \emph{Mathematical Programming}, 117\penalty0 (1-2):\penalty0 51--80,
  March 2009.

\bibitem[Tassa and Todorov(2010)]{tassastochastic}
Y~Tassa and E~Todorov.
\newblock Stochastic complementarity for local control of discontinuous
  dynamics.
\newblock In \emph{Robotics: Science and Systems}, pages 169--176, 2010.

\bibitem[Tedrake and the Drake Development~Team(2019)]{drake}
Russ Tedrake and the Drake Development~Team.
\newblock Drake: Model-based design and verification for robotics, 2019.
\newblock URL \url{https://drake.mit.edu}.

\bibitem[Hereid et~al.(2019)Hereid, Harib, Hartley, Gong, and
  Grizzle]{hereid2019rapid}
Ayonga Hereid, Omar Harib, Ross Hartley, Yukai Gong, and Jessy~W Grizzle.
\newblock Rapid trajectory optimization using c-frost with illustration on a
  cassie-series dynamic walking biped.
\newblock In \emph{2019 IEEE/RSJ International Conference on Intelligent Robots
  and Systems (IROS)}, pages 4722--4729. IEEE, 2019.

\bibitem[Posa et~al.(2016)Posa, Tobenkin, and Tedrake]{posa_stability_2016}
M.~Posa, M.~Tobenkin, and R.~Tedrake.
\newblock Stability {Analysis} and {Control} of {Rigid}-{Body} {Systems} {With}
  {Impacts} and {Friction}.
\newblock \emph{IEEE Transactions on Automatic Control}, 61\penalty0 (6), June
  2016.
\newblock ISSN 1558-2523.
\newblock \doi{10.1109/TAC.2015.2459151}.
\newblock Conference Name: IEEE Transactions on Automatic Control.

\bibitem[Hammoud et~al.(2021)Hammoud, Khadiv, and
  Righetti]{hammoud_impedance_2021}
Bilal Hammoud, Majid Khadiv, and Ludovic Righetti.
\newblock Impedance {Optimization} for {Uncertain} {Contact} {Interactions}
  {Through} {Risk} {Sensitive} {Optimal} {Control}.
\newblock \emph{IEEE Robotics and Automation Letters}, 6\penalty0 (3):\penalty0
  4766--4773, July 2021.
\newblock ISSN 2377-3766.
\newblock \doi{10.1109/LRA.2021.3068951}.

\end{thebibliography}






\end{document}